\documentclass[letterpaper, 10 pt, conference]{ieeeconf}

\IEEEoverridecommandlockouts                              % This command is only needed if 
\usepackage{multirow}
\usepackage{amsmath}
\usepackage{gensymb}
\usepackage{times}
\usepackage{helvet}
\usepackage{courier}
\usepackage{graphicx}
\usepackage{bm}
\usepackage{color}
\usepackage{epstopdf}
\usepackage{caption}
\usepackage{subcaption}
\let\labelindent\relax
\usepackage{enumitem}
\usepackage{calc}
\usepackage{multirow}
\usepackage{xspace}
\usepackage{booktabs}
\usepackage{mathrsfs}
\usepackage{array}
\usepackage{gensymb}
\usepackage{amsfonts}

\newcommand{\figref}[1]{Fig\onedot~\ref{#1}}
\newcommand{\equref}[1]{Eq\onedot~\eqref{#1}}
\newcommand{\secref}[1]{Sec\onedot~\ref{#1}}
\newcommand{\tabref}[1]{Tab\onedot~\ref{#1}}

\newcommand{\ve}[1]{{\mathbf #1}} % for displaying a vector or matrix
\newcommand{\hua}[1]{{\mathcal #1}}

\newcommand{\spa}[1]{{\mathbb #1}}

\DeclareRobustCommand\onedot{\futurelet\@let@token\@onedot}
\def\onedot{\ifx\@let@token.\else.\null\fi\xspace}
\def\eg{\emph{e.g.}}

\def\ie{\emph{i.e.}}

\def\wrt{w.r.t\onedot} 

\def\etal{\emph{et al.}}

\title{\LARGE \bf
Omnidirectional Depth Extension Networks
}

\author{Xinjing Cheng, Peng Wang$^\dagger$, Yanqi Zhou, Chenye Guan and Ruigang Yang% <-this % stops a space% <-this % stops a space
\thanks{Xinjing Cheng, Chenye Guan are with Robotics and Auto-driving Lab (RAL), Baidu Research, Baidu Inc., {\tt\small \{chengxinjing, guanchenye\}@baidu.com}. Peng Wang is currently working on Bytedance US AI Lab {\tt\small peng.wang@bytedance.com}. Ruigang Yang is with Kentucky university {\tt\small ryang@cs.uky.edu.}}%
\thanks{This work is done when Peng Wang, Yanqi Zhou, Ruigang Yang are employed at Baidu Research, Baidu Inc. }%
\thanks{$^\dagger$ denotes the corresponding author.}
}

\begin{document}
\maketitle
%\thispagestyle{empty}
%\pagestyle{empty}

%%%%%%%%%%%%%%%%%%%%%%%%%%%%%%%%%%%%%%%%%%%%%%%%%%%%%%%%%%%%%%%%%%%%%%%%%%%%%%%%
\begin{abstract}
%Depth estimation is critical for many applications, but most of efforts are put on regular pinhole cameras which are suffered from small field of view(FoV). 
% first make this task to be  very reasonable setup for robotic. 
% second propose our way of solving the task

Omnidirectional 360$^\circ$ camera proliferates rapidly for autonomous robots 
since it significantly enhances the perception ability by widening the field of view (FoV). However, corresponding 360$^\circ$ depth sensors, which are also critical for the perception system, are still difficult or expensive to have.
In this paper, we propose a low-cost 3D sensing system that combines an omnidirectional camera with a calibrated projective depth camera, where the depth from the limited FoV can be automatically extended to the rest of recorded omnidirectional image. 
To accurately recover the missing depths, we design an omnidirectional depth extension convolutional neural network (ODE-CNN), in which a spherical feature transform layer (SFTL) is embedded at the end of feature encoding layers, and a deformable convolutional spatial propagation network (D-CSPN) is appended at the end of feature decoding layers. The former re-samples the neighborhood of each pixel in the omnidirectional coordination to the projective coordination, which reduce the difficulty of feature learning, and the later automatically finds a proper context to well align the structures in the estimated depths via CNN \wrt the reference image, which significantly improves the visual quality.  % the geometric warping can be automatically learned in 
Finally, we demonstrate the effectiveness of proposed ODE-CNN over the popular 360D dataset, and show that ODE-CNN significantly outperforms (relatively 33\% reduction in depth error) other state-of-the-art (SoTA) methods.

%first solved the scale uncertainty of depth estimation from single image. 
%specifically, we extend the single view depth in cubemap to omnidirection with the help of rgb reference. 
%Besides, we introduce an feature transform layer which can convert the features learned by encoder from regular pinhole domain to omnidirectional domain. 
% Finnaly, we propose omnidirectional convolutional spatial propagation module to recover the structure details of omnidirecitonal depth, which is the extension of convolutional spatial propagation network designed for traditional depth completion task. 
\end{abstract}

\section{Introduction}
\label{sec:intro}

\begin{figure}[t]
\centering
\includegraphics[width=1.00\linewidth]{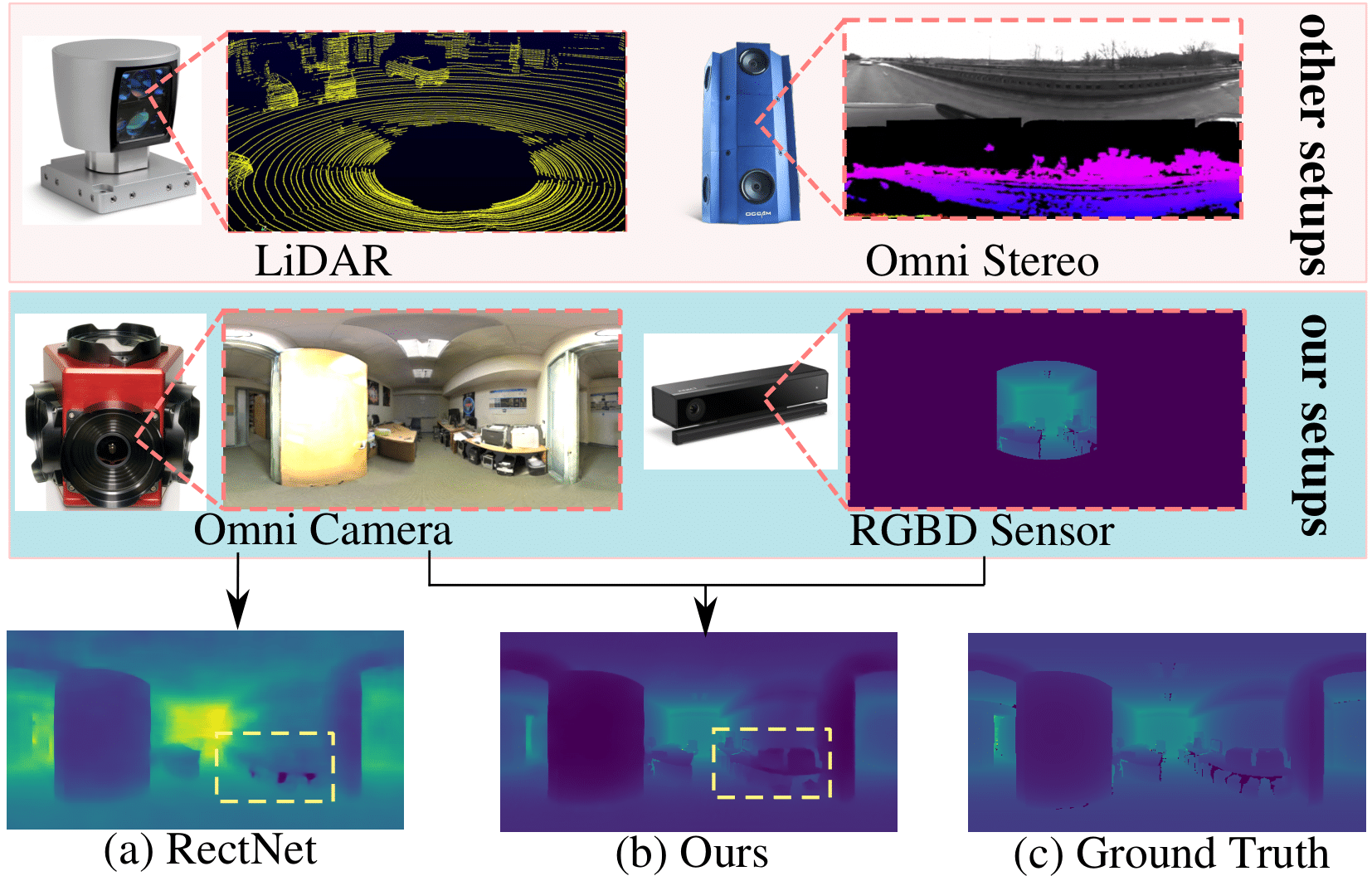}
\caption{Different from other setups which adopt LiDAR or omnidirectional stereo, we combine one projective depth sensor with omnidirectional camera and propose a new low-cost solution for omnidirectional depth. The last row shows that our ODE-CNN outperform SoTA RGB only OmniDepth network (RectNet)~\cite{360D} significantly both in depth accuracy and structural details.}
\label{fig:title_image}
\vspace{-1.0\baselineskip}
\end{figure}

% Nowadays, omnidirectional camera (OmniCamera) has become one of the most popular perception setups for indoor or outdoor robots, \eg~self-driving vehicles~\cite{chen2015deepdriving}, indoor robots~\cite{biswas2011depth}, thanks to its ability of observing 360\degree\ surrounding environment simultaneously. 
Ominidirection camera (OmniCamera) has been popular in robotics due to its 360 perception capability, which brings richer information to help with decisions.  
In real applications, besides RGB images, the per-pixel depth is also critical for many tasks, \eg~obstacle avoidance~\cite{fan2019getting}, 3D reconstruction~\cite{newcombe2011kinectfusion} and self-localization~\cite{zhang2014loam}. 
%The 360\degree\ field of view (FoV) helps the robots to act more reasonably based on much more complete information. 
% options for outdoor is LiDar and ToF kinect, structure light, stereo etc. 
% However, in real applications, it is not enough to only have RGB images, and understanding per-pixel 3D depth is also fundamental for obstacle avoidance~\cite{fan2019getting}, 3D reconstruction of the environment~\cite{newcombe2011kinectfusion} and self-localization~\cite{zhang2014loam}. 
While the cost for OmniCamera has dropped considerably, the cost for building omnidirectional depth sensors remains to be prohibitively high. 
For example, 
the popular depth sensors dependent on structure light, \eg~Kinect V1~\cite{zhang2012kinect1}, or time of flight (ToF), \eg~Kinect V2~\cite{lachat2015kinect2} are having limited FoV aroud 70\degree. Therefore, options of putting multiple Kinects together would require significant large bandwidth, yielding the difficulties of real-time synchronizing and calibration, and inducing significant latency in real applications. Another solution is installing a spinning LiDAR, \eg~Velodyne, while the cost is much higher compared with Kinect, and usually the frame rate is also significantly lower~\cite{velodyne} with only sparse depth points, as illustrated at the top in \figref{fig:title_image}. 

% the most popular device for autonomous driving (AD), spinning LiDAR, is more expensive than a typical vehicle, therefore preventing the wide dissemination of AD technology.  The alternative to combine multiple depth sensors to create an omnidirectional device suffers from the usual issues of calibration, synchronization, and bandwidth, in addition to cost. 

In this paper, rather than working on  the  side  of hardware, motivated by the high quality single view depth estimation from  deep fully convolutional networks~\cite{huang2017densenet,wang2016surge}, we propose a software-based solution to alleviate the issue. Nevertheless, directly applying depth learned from a single  omnidirectional image (OmniImage) may suffer from the scale confusion when switching environments as illustrated in \figref{fig:title_image}(a). 
Therefore, in this work, we present a hybrid setup that combines a regular omnidirectional video camera with a regular FoV depth senor, \eg~a Kinect Camera.
The depth camera provides metric depth values for part of the scene,
and the depth values for the remaining regions are extended based on both the known depths and the image content. Compared to single-image based method~\cite{360D}, our hybrid depth sensor is much accurate as shown in \figref{fig:title_image}(b). 

% we propose to first adopt one depth sensor with a limited FoV, and then extend the perceived accurate depth to the rest of image, which generates much more accurate depth estimation results and yields significant better trade-off between depth perceiving speed and accuracy, which is .
%This technique becomes more applicable with the emerge of  which produces high quality single view depth estimation results. 
% talk about it is possible to do this with deep network, and the limitations on our problem. 

Practically, building a depth CNN performing over OmniImage is different from that over regular projective images, since the warped image texture under spherical coordinate does not suitable to use the feature learned under projective coordinates.
Therefore, researchers mainly adopt two ways to tackle this. The first one is transforming an omnidirectional image (OmniImage) to six cube maps, and estimating depth for each cube map~\cite{liu2015cubemap}. However, the discontinuities/non-smoothness of depths can be significant at the boundary when stitching different depth maps. 
The second option is adopting the spherical convolution~\cite{coors2018spherenet,360D} over the OmniImage directly, where the convolutional neighborhoods of each pixel are re-sampled dynamically \wrt equirectangular coordinate. %as shown in \figref{}.
However, the practical latency of the modified convolution as proposed in~\cite{coors2018spherenet} is significantly slower than a normal convolution, which is not practice when performed at every layer of CNN over the OmniImage with high resolution. Therefore, we chose to allocate the modified convolution at the end of encoding, where the spatial resolution of the convolutional feature has been reduced significantly (usually 16$\times$ or 32$\times$ after encoding). We call this operation as spherical feature transform layer (SFTL), which reduces the difficulty of feature learning yielding better accuracy. Besides, this also inspires us to propose a more generalized transformation by allowing its convolutional neighborhoods of a pixel to be automatically learned \wrt the dataset. In practice, we induce deformable conv~\cite{dai2017deformable} to act inside SFTL, yielding even better results. % taking geometry transformation to eliminate distortion between equirectangular and pinhole projection as proposed in~\cite{coors2018spherenet}, 
% we additionally introduce  , which learns the convolutional neighborhoods directly from data, 

Moreover, depths obtained directly from a CNN have less focuses on image details and structures, yielding non-satisfied estimation. 
%as illustrated in \figref{fig:title_image}(c). 
Therefore, we borrow an efficient depth refinement strategies, \ie~ convolutional spatial propagation network (CSPN)~\cite{cheng2018cspn}, to handle the issue.
Here, similarly, we modify the sampled neighborhood of CSPN \wrt equirectangular coordinates using the inverse gnomonic projection (IG)~\cite{frederick2018map}, namely IG-CSPN, to enhance its performance. In addition, we also choose to deform its convolutional neighborhoods similiar as SFTL, namely Deformable CSPN (D-CSPN). We find the D-CSPN adapts well with both image context and distortion, yielding significant performance boost over CSPN and IG-CSPN. 
Finally, after properly embed the two proposed modules, we illustrate our whole framework for ominidirectional depth extension (ODE) in \figref{fig:framework},  which is named as ODE-CNN.

%and equirectangular projections. 
% Conversely, the equirectangular projections preserve the spatial relationship of contents which is more suitable for CNN-based method but suffer from heavy distortions in the polar regions due to the horizontal and vertical grid coordinates are mapped from latitude and longitude of omnidirectional images. 
% To address the heavy distortion, 

To validate our proposed depth extension setup, we perform various study over the recently proposed dataset with omnidirectional images~\cite{360D}, where we show firstly by adopting additional one depth senor with limited FoV can significantly reduce overall depth error from the estimation from a single RGB image, and jointly adopting our ODE-CNN achieve relatively 33\% error reduction vs. the previous SoTA networks~\cite{360D}.  

In summary, our paper has following contributions:
\begin{enumerate}
\item We propose a low-cost omnidirectional depth sensing system by combining an OmniCamera with a regular depth sensor (limited FoV), which achieves strong depth sensing efficiency (50ms) and accuracy (4\% relative depth error).
\item For building the depth network, we propose to use a spherical feature transform layer at the end of encoder to reduce the difficulty of feature learning, and extend the CSPN~\cite{cheng2018cspn} to inverse gnomonic projection CSPN (IG-CSPN) and deformable CSPN (D-CSPN), which better recover the structural details of estimated depths.
\end{enumerate}

% We will release codes upon the acceptance of this article. 
\vspace{-0.5\baselineskip}
\section{related work}
\vspace{-0.\baselineskip}
% To the best of the author's knowledge, there are countable deep neural network architectures which are designed for omnidirectional inputs, the omnidirectional dense depth estimation networks are even fewer. 
% To Besides this, we are the first to work on the omnidirectional dense depth extension with sparse depth inputs. So, in this section, we review the most related methods.
As discussed in \secref{sec:intro}, with the rapid development of deep neural networks~\cite{he2016resnet,huang2017densenet} and growth accessibility for large scale depth estimation dataset~\cite{silberman2012nyu,geiger2013kitti}. Depth estimation from CNNs has largely outperform the traditional ones~\cite{hirschmuller2005sgm}. In this section, we majorly review relevant works about the dense prediction with OmniImages, and elaborate their relationship with ODE-CNN proposed in this paper.

\subsection{Single view depth estimation and completion}
The idea of dense depth estimation with CNNs is first introduced in~\cite{eigen2014depth}, and then the quality of estimation is vastly improved by jointly adopting conditional random field (CRF)~\cite{liu2015depthcrf}, combining with auxiliary tasks, such as semantic~\cite{wang2016surge,wang2015towards}, normal~\cite{zhang2019pattern}, edges~\cite{song2018edgestereo,yang2018lego} and especially with the development of fully convolutional networks (FCN)~\cite{long2015fully} with stronger CNN backbones such as ResNet~\cite{he2016resnet}. 

Although the estimated depth map from a single image achieves impressive quality, it still suffers from instability of domain transfer, \ie~performing the learned model on other unseen places. To alleviate the issue, most recently, some researchers propose to combine sparse depths captured through devices like LiDAR, and perform depth completion~\cite{cheng2018cspn}, while others propose to combine some low cost depth sensors, where dense depths are partially captured, and perform depth enhancement~\cite{matsuo2015depthenhance} or in-painting~\cite{miao2012inpainting}. In both setting, depth estimation accuracy and generalization ability are significant improved.

In this work, since we are dealing with the OmniImages whose FoV is much larger than that of a single view image, we here propose a novel task called omnidirectional depth extension which extends the dense depths captured at front view to the rest of corresponding OmniImage. As discussed in \secref{sec:intro}, it is a more practical setting than depth completion or in-painting. 
% The single view depth estimation gets more popular for a long time. 
% There are numerous algorithms developed though supervised learning(), semi-supervised methods() and unsupervised learning() and made significant progress. 

% Depth completion is the task which obtain per-pixel depth from sparse depth with reference of RGB image.

\begin{figure*}[htbp]
\centering
\includegraphics[width=1.00\linewidth]{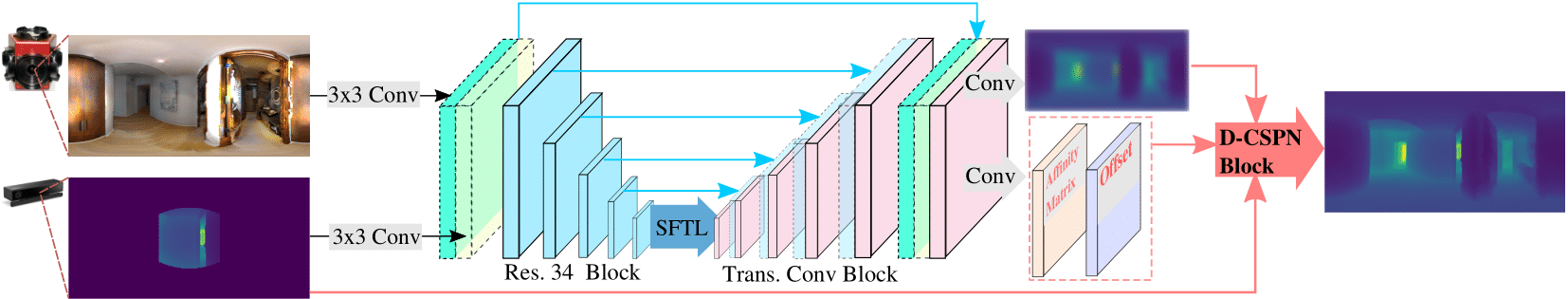}
\caption{Network architecture of our ODE-CNN(best view in color). We introduce spherical feature transform layer (SFTL) to transform the features from pinhole model to equirectangular model~\secref{sec:ftl}. At the end of the network, we generate the affinity matrix and convolutional offsets for deformable convolutional spatial propagation network (D-CSPN), which refine the predicted depth with more detailed structures. }
\label{fig:framework}
\vspace{-1.4\baselineskip}
\end{figure*}

\subsection{Deep network over OmniImage}
% single view omni prediciton & multiple view setup 
To deal with the projection distortion on OmniImages when processing with CNNs, researchers propose various technologies either modifying over the input image or over the convolutional kernels. 
For the former strategy, early works~\cite{xiao2012recognizing,su2017making} propose to process repeatedly over each projected image around the sphere, and merge the processed results, yielding high computational cost in real applications.
Later, works such as estimating saliency~\cite{monroy2018salnet360} or artistic style transfer with OmniImages~\cite{ruder2018artistic} propose to project the OmniImage to faces of a cube (cubemap), and then processing each cubemap independently with regular CNN. However, such a solution is not recommended for depth estimation, since there would be significant depth artifacts along the boundary connecting different estimated results. In addition, for depth extension in our case, processing each views independently will ignore the critical information from the depth sensor.

Therefore, we choose the methodology of processing the OmniImage as a whole by modifying the convolutional kernels. In this direction, Su \etal~\cite{su2017learning} first propose to learn convolution weights for equirectangular projected OmniImages by transferring them from an pre-trained network from 2D images with a pinhole camera. 
Cohen \etal~\cite{cohen2017convolutional,cohen2018spherical} propose spherical CNNs that are based on a rotation equivalent definition of spherical cross-correlation, which is specially designed for classification tasks. %Khasanova \etal~\cite{} propose to adopt graph-based representation . 
Later works~\cite{khasanova2017graph,coors2018spherenet,tateno2018distortion} adopt the intrinsic relationship between sphere and equirectangular coordinates, \ie~ inverse gnomonic projection, to re-sample the neighborhoods of convolution according to a spherical tangent plane. %  can be  adopt inverse gnomonic projection to get where the convolutional kernel located in omnidirectional image. 
Most recently, in mapped convolution~\cite{Eder2019mapped}, Eder \etal~ indicate that tangent plane is not the optimal model for OmniImage, and they induce an inverse equirectangular projection for the convolutional kernels to better compensate the distortion. Finally, in our perspective, the spherical convolution can be seen as a special case of free-form deformable convolution as introduced in some recent works~\cite{jeon2017active,dai2017deformable} for object recognition. % the shape of convolutional kernels is adaptively changed \wrt to image content without considering the prior knowledge of image distortion. 
However, it is only adopted in segmentation of fish-eye images for now~\cite{deng2018restricted}.  

In our proposed ODE-CNN for handling depth estimation, we consider both the underlining formula of equirectangular projection, and free-form deformation jointly inside the architecture. Thanks to the recently released large scale OmniDepth dataset~\cite{360D}, we are able to learn inspiring deformed kernel shapes and features from scratch, yielding significant improvement. 
%  over the SoTA strategies~\cite{360D} in this task
Last but not the least, we would like to note that there are other options of acquiring omnidirectional depth map such as spherical stereo estimation by setting up multiple OmniCameras~\cite{lin2018real}, which could be served as complementary components in our framework. 

\section{Approach}
% \vspace{-0.1\baselineskip}
% discribe the set up ? architecture, and then discribe two detailed components. 

Given the setup described in~\secref{sec:intro}, we here elaborate our designed encoder-decoder architecture, and the two proposed modules, \ie~SFTL and D-CSPN, for enhancing the performance over the depth extension task. 
%Then we describe the feature transformed layer which convert the feature space from perceptive to equirectangular. 
% Finally, we elaborate the panoramic convolutional spatial propagation (PCSP) module we proposed, which is an anisotropic diffusion process and the diffusion tensor is learned from a deep CNN directly from the given image and sparse depth.

\subsection{Network Architecture}
% We formulate the panoramic depth estimation problem as a deep regression learning problem. 
As displayed in~\figref{fig:framework}, our network architecture follows an encoder-decoder paradigm proposed by~\cite{ma2018self}, which is a improved ResNet-34~\cite{he2016resnet} by adding mirror skip connections at corresponding layers with same spatial resolution from encoder to decoder.

The input OmniImage and the partial dense depth map, when available, are separately processed by their initial convolutions. The convolved outputs are concatenated into a single tensor, which acts as input to the encoder. 
The encoder consists of five residual blocks which downsample the spatial 
resolution of the feature map 16$\times$ \wrt the image resolution. 
On the other side, the decoder has the reverse structure, where transposed convolution layers are adopted here to upsample the feature map to the original resolution for dense depth estimation.  In the network, all convolutional layers are followed by batch normalization~\cite{ioffe2015bn} and ReLU~\cite{nair2010relu}.

To alleviate the distortion problem of OmniImage, at the end of encoder, we adopt a SFTL
%as introduced in SphereNet~\cite{coors2018spherenet} 
to obtain spherical neighborhood in the equirectangular image with inverse gnomonic projection or deformable convolution. 
In addition, to recover the structure detail of the estimated depth map, at the end of decoder, we adopt the module of CSPN~\cite{cheng2018cspn}, and modify its propagation neighborhood to be dynamically changing \wrt to the pixel location. In the following sections, we elaborate the difference of both modules.

% which lift local CNN operations from the regular image domain to the sphere surface, 
% we propose a feature transform layer (FTL) located between encoder and decoder, they key idea of FTL is convert the feature from regular image domain to equirectangular domain.

% The input color image and sparse depth (if available) are processed separately by their initial convolutions, convolution outputs are concatenated into one tensor, which acts as the inputs to the Encoder-FTL-Decoder block. 
% Finally, the output is processed by our panoramic convolutional spatial propagation(PCSP) module, which is a refinement module which can train by a end-to-end faction within the whole network. The final output will be the same resolution as network input. 

\subsection{Spherical feature transform layer (SFTL)}
\label{sec:ftl}
% A spatial feature transform considering mapping 
In this section, we elaborate the concept of spatial feature transform (SFT) for convolution, and then introduce our strategy of learning per-pixel transform for OmniDepth estimation. 
Formally, suppose a feature map ${\ve{H}^l \in \spa{R}^{c\times h\times w}}$ is from the image represented at the target coordinate, \ie an OmniImage. Here, ${h, w, c}$ are height, width and number of feature channels respectively, and $l$ is the layer id.
Here, given coordination transform $g_{\ve{x}_t}()$ at a target location $\ve{x}_t$ from a source coordinate, \eg~a tangent plane, we may map any points at source coordinate $\ve{x}$ to target by $g_{\ve{x}_t}(\ve{x})$.
When performing convolution over the OmniImage, we want to sample neighborhoods from the source planar surface without distortion. Therefore, the convolution needs to use the transformed spatial neighborhoods, which can be written as,
\begin{align}
    \ve{H^{l+1}}(\ve{x}_t) = \sum_{\Delta{\ve{x}_s} \in \hua{N}(\ve{x}_s)} \kappa(\Delta{\ve{x}_s}) * \ve{H^l}(g_{\ve{x}_t}(\ve{x}_s + \Delta{\ve{x}_s}))
\label{equ:sftl}
\end{align}
where $\kappa(\Delta_{x})$ represents the convolutional kernel weights of shape $c_o \times c$, and $*$ is matrix multiplication. Here, we set $g_{\ve{x}_t}(\ve{x}_s) = \ve{x}_t$ for making $\ve{x}_t$ as the convolutional center, and $\hua{N}(\ve{x}_s)$ is the set of locations around $\ve{x}_s$ at source coordination for convolution. For example, $\Delta{\ve{x}_s} \in [(-1, -1), \cdots, (1, 1)]$ represents a $3\times 3$ convolutional kernel that is popularly adopted in many SoTA networks~\cite{he2016resnet}.
$\ve{H}_t(\ve{x}_t)$ indicates the feature representation at $\ve{x}_t$, and a bilinear interpolation is usually performed for fractional locations after tranformation $g_{\ve{x}_t}()$ .

In spherical transformation~\cite{coors2018spherenet, tateno2018distortion, cohen2018spherical}, the source to target transformation $g_{\ve{x}_t}(\ve{x})$ is called inverse gnomonic projection~\cite{frederick2018map}. 
Here, the target coordination is over a sphere, and given a point $\ve{x}_t=(\phi, \theta)$ in sphere surface, the pixels in OmniImage is uniformly sampled \wrt the latitude coordinate $\phi \in \left [-\frac{\pi }{2}, \frac{\pi }{2} \right ]$   and the longitude coordinate $\theta \in \left[-\pi, \pi \right]$.  
For each point $\ve{x}_t$, the source coordinate is over the local tangent plane $ \ve{x}_s=(x, y)$ centered at $\ve{x}_t$. 
Then, the inverse gnomonic projection function $g_{\ve{x}_t}(\ve{x}_s)$ can be formulated as follow,
 %~\cite{coors2018spherenet} and ~\cite{tateno2018distortion} address this domain gap by  inverse gnomonic projection, 
% which is the distortion introduced by different projection models. 
\begin{align}
    g_{\ve{x}_t}(\ve{x}_s) &= (\phi(\ve{x}_s), \theta(\ve{x}_s)) \nonumber \\
    \phi(\ve{x}_s) &= sin^{-1}(cos\varphi \cdot sin\tau_\phi + \frac{y \cdot sin\varphi \cdot cos\tau_\phi}{\rho }) \\
    \theta(\ve{x}_s) &= \tau_\theta + tan^{-1}(\frac{x\cdot sin\varphi }{\rho \cdot cos\tau_\phi\cdot cos\varphi - y\cdot sin\tau_\phi\cdot sin\varphi}) \nonumber 
\label{equ:igp}
\end{align}
% where the $T$ here is a domain transform function. This can be seen as a domain adaptation problem under fixed geometric transformation constraint. 
% The first insight is to use gnomonic projection to eliminate the fixed gap. 
where $\rho =\|\ve{x}_s\|_2 = \sqrt{x^2+y^2}$, $\varphi= tan^{-1}\rho$. Here, to perform convolution, we follow \cite{coors2018spherenet} to sample neighborhood $\Delta{\ve{x}_s}$ on the tangent plane.

% The gnomonic projection is a effective way to transform feature to sphere space. However, it maps the sphere onto a tangent plane where our target is cylinder plane, which should be mapped by equirectangular projection, \ie, the $y$ coordinate in the image is related to the latitude $\phi$. 
However, in real cases, it might be not optimal to model the transform layer as a fixed geometric transformation~\cite{dai2017deformable}, a feature can be much stronger when image context is also embedded for the transformation. Therefore, in this work, we propose to learn a deform offset inside the tangent plane centered at $\ve{x}_t$, \ie~setting $\Delta{\ve{x}_s}$ in \equref{equ:sftl} as a trainable variable rather than fixed neighborhood. 
Specifically, at the end of encoder, the network outputs additional $\Delta{\ve{x}_s}$ using a $3\times 3$ convolutional layer, which is then embedded for deformation in the source coordinate of tangent planar before the transformation.
% we act it as a learning progress which learn the transformation directly by a convolutional layer, shown as below:
% \begin{align}
%     \ve{H_e(x,y)} = \sum\nolimits_{a,b = -(k-1)/2}^{(k-1)/2} \omega [\ve{P}(a,b)]\odot \ve{H_p}(x-a, y-b)
% \end{align}
% where $k$ is the kernel size we adopt and $\ve{P} \in \spa{R}^{ k^2 \times 2}$ is the learned transformation by convolution on \ve{H_p}.

\textit{Complexity analysis.} We adopt the CUDA implementation of deformable convolution~\cite{dai2017deformable} for SFTL, where the key component is \texttt{deform-im2col} function. Compared with a regular convolution implemented in ~\cite{he2016resnet}, where the \texttt{im2col} function inside can be easily implemented by shifting the image features. SFTL has the extra cost of predicting per-pixel spatial transform offset, dynamically indexing image features and bilinear interpolation. Therefore, our proposed SFTL has extra memory cost of $O(hwk^2)$, where $k$ is the size of kernel, and speed latency of $O(k^2)$. In practice, when performing over a Nvidia P40 GPU, SFTL is 30\% slower and requires $(1 + 2*k^2/c)\times$ memory comparing to a regular convolution, hindering the deployment of the module to every convolutional layers. Therefore, to minimize the extra cost in the network, we put SFTL at the end of encoder where spatial resolution of image feature is $16\times$ reduced.

\begin{table*}[t]
\centering
\caption{Quantitative results for panoramic depth estimation experiments on 360D~\cite{360D} datasets. }
\label{tbl:quanti_res}
\fontsize{10}{10}\selectfont
\bgroup
\def\arraystretch{1.3}
\setlength{\tabcolsep}{6.0pt} % General space between cols (6pt standard)
\begin{tabular}{l|c|c|c|cccc|ccc} 
\hline
\multirow{2}{*}{Network}        & \multirow{2}{*}{SFT} & \multirow{2}{*}{CSPN} & \multirow{2}{*}{PD} & \multicolumn{4}{c|}{Lower the Better} & \multicolumn{3}{c}{Higher the Better}  \\ 
\cline{5-11}&                     &                       &                     & Abs Rel~ & Sq Rel~ & RMSE   & RMSLog  & $\delta_{1.25}$  & $\delta_{1.25^2}$ & $\delta_{1.25^3}$              \\ 
\hline
UResNet~\cite{360D}                         & -                   & -                     & -                   & 0.0835   & 0.0416  & 0.3374 & 0.1204  & 93.19 & 98.89   & 99.68                \\ 
\hline
RectNet~\cite{360D}                         & -                   & -                     & -                   & 0.0702   & 0.0297  & 0.2911 & 0.1017  & 95.74 & 99.33   & 99.79                \\ 
\hline
\multirow{8}{*}{ODE-CNN } & -                   & -                     & -                   & 0.0642   & 0.0174  & 0.2086 & 0.0964  & 96.86 & 99.59   & 99.88                \\
                                & -                   & -                     & front               & 0.0624   & 0.0173  & 0.2090 & 0.0943  & 97.03 & 99.61   & 99.87                \\
                                & IGT        & -                     & front               & 0.0521   & 0.0129  & 0.1794 & 0.0828  & 97.91 & 99.68   & 99.90                \\
                                & DIGT           & -                     & front               &0.0494    &0.0126   &0.1780 &0.0808   &98.06  &99.66    &99.89                 \\
                                & -                   & CSPN~\cite{cheng2018cspn}                  & front      & 0.0539   & 0.0140  & 0.1849 & 0.0850  & 97.57 & 99.59   & 99.88                \\
                                & -                   & IG-CSPN                 & front               & 0.0523   & 0.0139  & 0.1812 & 0.0857  & 97.59 & 99.64   & 99.89                \\
                                & -                   & D-CSPN                & front               & 0.0511   & 0.0139  & 0.1778 & 0.0850  & 97.61 & 99.65   & 99.89                \\
                                & DIGT  & D-CSPN     & front      &\textbf{0.0467} &\textbf{0.0124} &\textbf{0.1728} &\textbf{0.0793}         &\textbf{98.14} &\textbf{99.67} &\textbf{99.89}                      \\
\hline
\end{tabular}
\vspace{-1.0\baselineskip}
\egroup
\end{table*}

% f $O($Since perform
\subsection{Deformable convolutional spatial propagation network}
\label{sec:d-cspn}

\begin{figure}[t]
\centering
\includegraphics[width=1.00\linewidth]{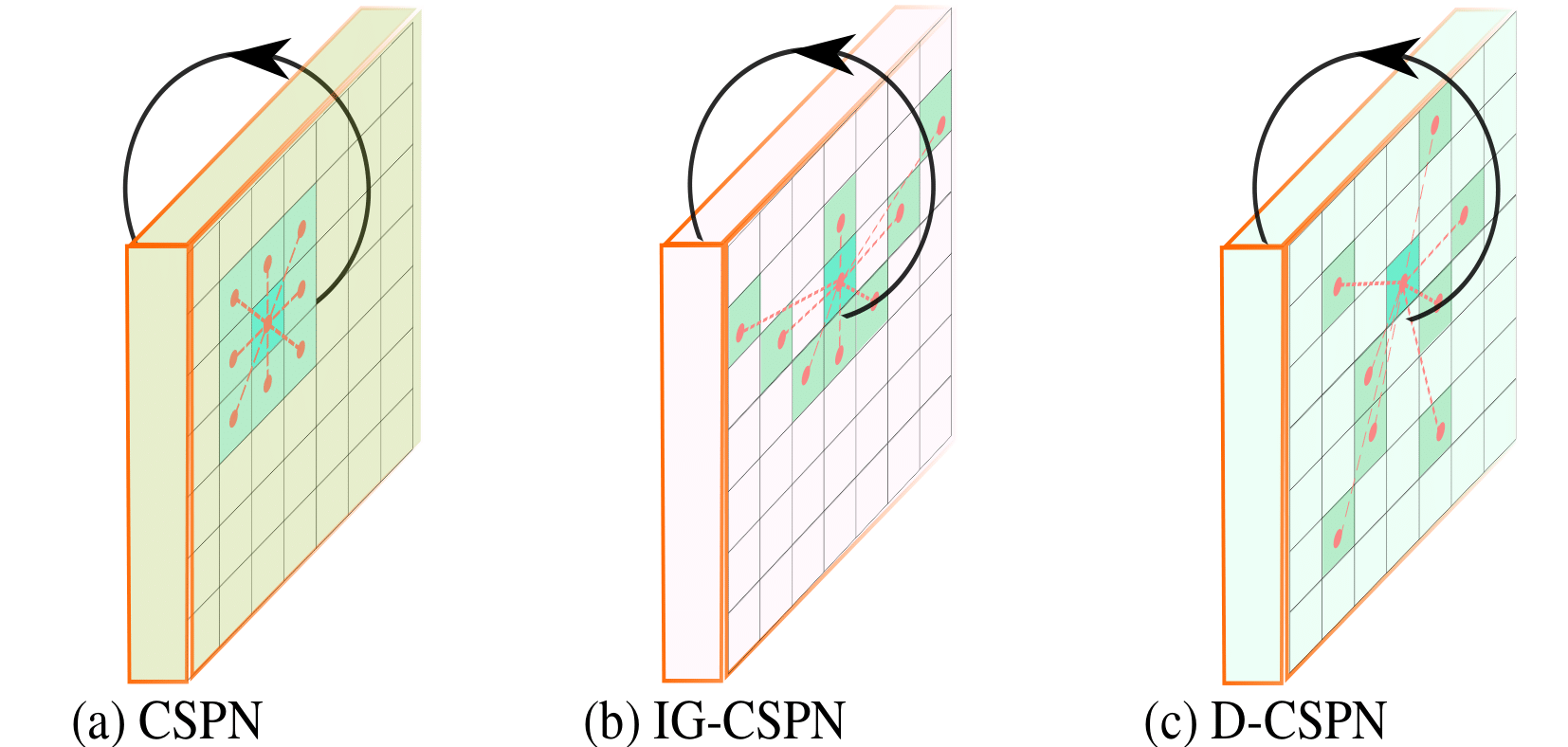}
\caption{(a) Convolutional spatial propagtion network (CSPN). (b) Inverse gnomonic CSPN (IG-CSPN) with neighborhood sampled with predefined formula in \equref{equ:igp}. (c) Deformable CSPN with neighborhood automatically learned based on inverse gnomonic projection. }
\label{fig:cspn}
\vspace{-1.0\baselineskip}
\end{figure}

As illustrated in \figref{fig:framework}, the estimated OmniDepth directly after the process of decoder is still blurry due to it has less focus on edges and detailed structures in the OmniImage. 
Therefore, we adopt CSPN to refine the final output. Specifically, as illustrated in \figref{fig:cspn}, CSPN has a similar computational formula with convolution, where the differences are first it adopts a per-pixel transformation kernel to substitute the kernel weights, and second a recurrent operation is performed inside for propagation. Here, we also embed the spatial transform as induced in~\equref{equ:sftl} when performing the convolution.

Formally, given the feature blob $\ve{H}_{\tau} \in \spa{R}^{c\times h\times w}$ at time step $\tau$, one step CSPN~\cite{cheng2018learning} at location $\ve{x}_t$ in our case can be written as, 
\begin{align}
    \ve{H}_{\tau + 1}&(\ve{x}_t) = \kappa_{\ve{x}_t}(\ve{0}) \odot \ve{H}_{0}(g_{\ve{x}_t}(\ve{x}_s) +  \\ 
    &\sum_{\Delta{\ve{x}_s} \in \hua{N}_k(\ve{x}_s)} \kappa_{\ve{x}_t}(\Delta{\ve{x}_s}) \odot \ve{H_{\tau}}(g_{\ve{x}_t}(\ve{x}_s + \Delta{\ve{x}_s})) \nonumber \\
\mbox{where,~~~~}
    \kappa_{\ve{x}_t}&(\Delta{\ve{x}_s}) = \hat{\kappa}_{\ve{x}_t}(\Delta{\ve{x}_s}) / {\sum_{\Delta{\ve{x}_s} \in \hua{N}_k} |\hat{\kappa}_{\ve{x}_t}(\Delta{\ve{x}_s})|}, \nonumber \\
    \kappa_{\ve{x}_t}&(\ve{0}) = \ve{1} - \sum\nolimits_{\Delta{\ve{x}_s} \in \hua{N}}\kappa_{\ve{x}_t}(\Delta{\ve{x}_s})  \nonumber
\label{eqn:dcspn}
\end{align}
where $\odot$ is the element-wise production between the transformation kernel and image feature. The kernel $\kappa_{\ve{x}_t}$ is location dependent with shape of $c \times k \times k$, and $\hua{N}_k(\ve{x}_t)$ is the neighborhood pixels of $\ve{x}_t$ with a $k\times k$ kernel. The affinities output from a network $\hat{\kappa}_{\ve{x}_t}()$ are properly normalized which guarantees the stability of CSPN, and the whole process will iterate $N$ times to obtain the final results. Here, $k, N$ are predefined hyper-parameters.

Similar with SFTL, as illustrated in \figref{fig:cspn}, we also adopt the inverse gnomonic (IG) projection (as stated in \equref{equ:igp}) to re-sample the neighborhood for CSPN, which we call IG-CSPN.  Then, we allow the offset $\Delta{\ve{x}_s}$ to be a trainable variable predicted from a shared network as illustrated in \figref{fig:framework}, which we call deformable CSPN (D-CSPN). In our experiments (\secref{sec:exp}), we show IG-CSPN is more effective than the vanilla CSPN~\cite{cheng2018learning} with the same configuration, and D-CSPN achieves even better performance, which is adopted as our final module for depth refinement.

% \begin{align}
%     \ve{H}_{\tau + 1} = &\sum\nolimits_{a,b = -(k-1)/2}^{(k-1)/2} \kappa_{i,j}(\ve{P}(a,b)) \odot \ve{H}_{i-a, j-b, t} \nonumber \\
% \mbox{where,~~~~}
%     &\kappa_{i,j}(\ve{P}(a,b)) = \frac{\hat{\kappa}_{i,j}(\ve{P}(a,b))}{\sum_{a,b, a, b \neq 0} |\hat{\kappa}_{i,j}(\ve{P}(a,b))|}, \nonumber\\
%     &\kappa_{i,j}(0, 0) = 1 - \sum\nolimits_{a,b, a, b \neq 0}\kappa_{i,j}(\ve{P}(a,b))
% \label{eqn:pcsp}
% \end{align}
% Given a depth map $D_o \in \spa{R}^{m\times n}$ that is output from decoder, and image $\ve{X} \in \spa{R}^{m\times n}$, ~\cite{cheng2018cspn} refine the depth map to a new depth map $D_n$ within $N$ iteration steps, which first reveals more details of the image, and second improves the per-pixel depth estimation results. However, ~\cite{cheng2018cspn} is proposed on perceptive image. We extend the cspn to panoramic image and propose panoramic convolutional spatial propagation(PCSP) module, which defined as Eq.~\ref{eqn:pcsp}

Finally, to preserve the accurate depths from the partial dense depth map obtained by the depth sensor,  same with CSPN for depth completion, we also induce a replacement step. Specifically, let $\ve{H}^p$ to be a hidden representation for the partial dense depth map $D^p = \{d_{\ve{x}}^p\}$, and the replacement can be written as,
\begin{align}
    \ve{H}_{\tau + 1}(\ve{x}_t) = (1 - m_{\ve{x}_t}) \ve{H}_{\tau + 1}(\ve{x}_t)  +  m_{\ve{x}_t} \ve{H}^p(\ve{x}_t)
\end{align}
where $m_{\ve{x}_t} = \spa{I}(d_{\ve{x}_t}^p > 0)$ is an indicator for the availability of partial depth map. Thanks to D-CSPN, the generated depths produce much better details align with structures in the OmniImage, and the transition between the depths from the depth sensor and the estimated depths from ODE-CNN is smooth and unnoticeable.

%\paragraph{Deform with inverse gnomonic projection.}

% is a direct extend

% \paragraph{Deform with freeform projection.}

% Similar as Sec.~\ref{sec:ftl}, we proposed two way to get the transformation $\ve{P}$, \ie, inverse gnomonic projection and learn from feature directly. At each step of propagation, we replace the iterated output with ground-truth sparse depth $D_s$, shown in Eq.~\ref{eqn:srp}. 

\vspace{-0.3\baselineskip}
\section{Experiments}
% \vspace{-0.\baselineskip}
\label{sec:exp}

In this section, we validate ODE-CNN exhaustively, and in the following, we describe our experimental setting, \ie the datasets, metrics and implementation details. 
Then, results from a detailed ablation study of each module we proposed in ODE-CNN are presented.
Finally, we compare against other SoTA methods, and qualitatively illustrate the improvements in our estimated OmniDepth maps.

\begin{figure*}[t]
\centering
\includegraphics[width=1.00\linewidth]{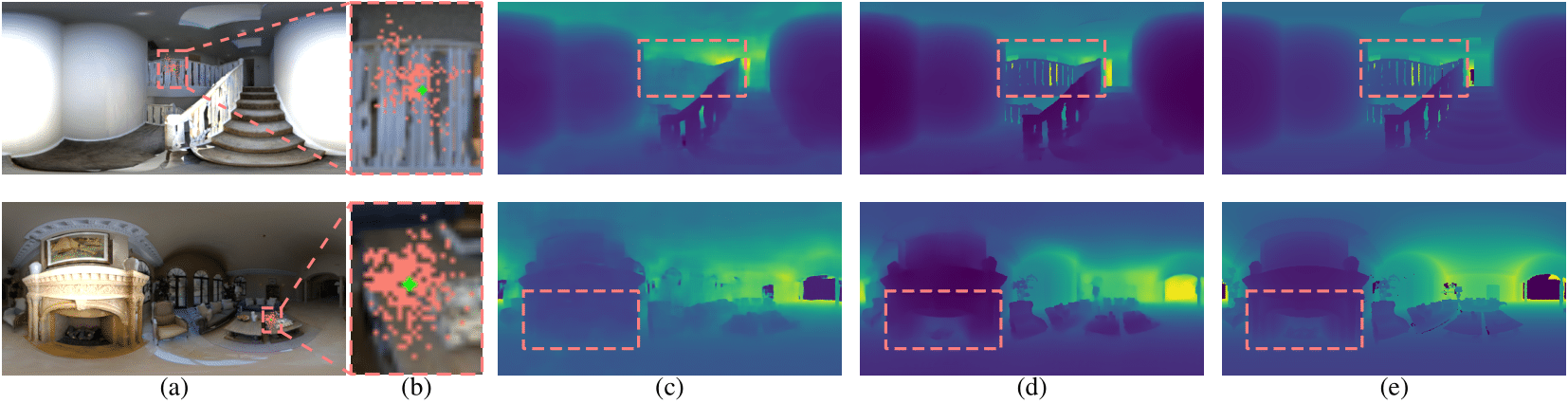}
\caption{Qualitative comparison with RectNet. (a) Image, (b) learned deformable offset at particular point, (c) results from RectNet~\cite{360D}, (d) results from ODE-CNN, (e) ground truth.}
\label{fig:quali}
\vspace{-1.0\baselineskip}
\end{figure*}

\begin{figure}[t]
\centering
\includegraphics[width=1.00\linewidth]{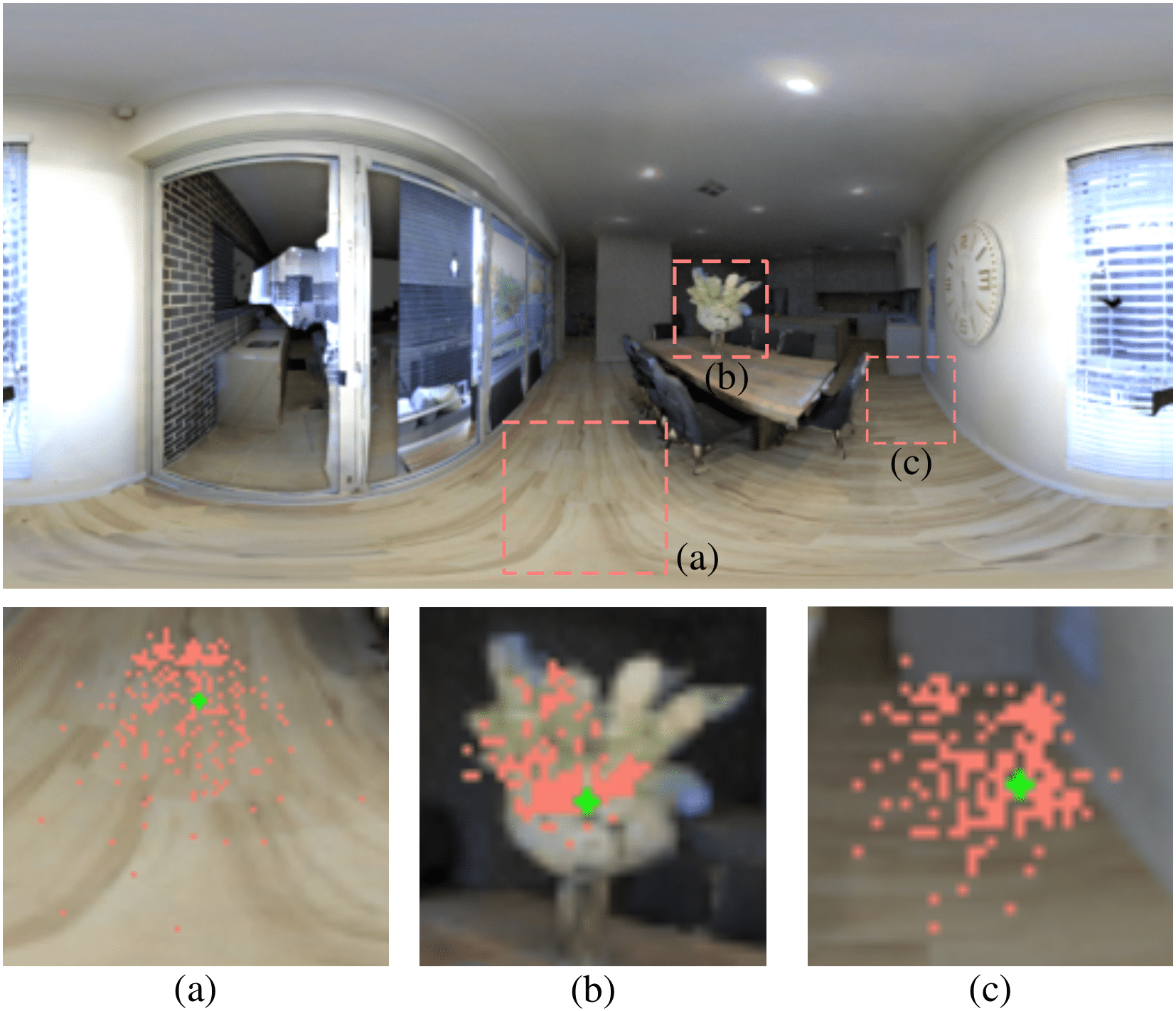}
\caption{Learned contexts though our deformable convolutional spatial propagation network, we zoom in three different regions, where show that our method will consider both distortion and context during propagation.}
\label{fig:offset}
\vspace{-1.0\baselineskip}
\end{figure}

\subsection{Experimental setting}
% \addlinespace
\textit{DataSet.} The 360D~\cite{360D} is a large-scale indoor spherical RGB-D dataset, which aggregates two realistic datasets, \ie, Standford 2D-3D~\cite{standford2d3d} and Matterport 3D~\cite{chang2017matterport3d}, and two computer generated (CG) dataset, \ie, SunCG~\cite{sunCG} and SceneNet~\cite{dai2017scannet}. 
Here, we adopt the same data split strategy as~\cite{360D}. Specifically, the train and test split are set as,
\begin{enumerate}
\setlength{\itemsep}{5pt}
\item Training: remove scenes which contain regions with very large and small depth value in the remaining data in the Standford 2D-3D, Matterport 3D and SunCG, containing 34679 samples.
% \item Validation: use the whole SceneNet as validation for tuning our hyper-parameters.
\item Testing: remove one complete area from Standford 2D-3D, and three complete buildings from Matterport 3D and three CAD scenes from SunCG，containing 1298 samples totally.
\end{enumerate}

\textit{Metrics.} 
We adopt the same metrics as proposed in OmniDepth~\cite{360D}, including absolute/square relative error (Abs/Sq Rel), rooted mean square error (RMSE),  RMSE in log space(RMSLog) and $\delta_t$, where $\delta_t$ means $\%$ of correct predicted depth $d \in D$ under different criterion, \ie~$max(\frac{d^*}{d}, \frac{d}{d^*})<t$, where $t \in \{1.25, 1.25^2, 1.25^3\}$. 
We refer readers to the original paper for the detailed formula of these metrics due to space limits.
% We adopt the same metrics as \cite{360D}. Given ground truth depth $D^* = \{d^*\}$ and predicted depth $D = \{d\}$, the metrics include:  
% \begin{enumerate}
% \setlength{\itemsep}{5pt}
% \item \textbf{Abs Rel}: $\frac{1}{|D|}\sum_{d \in D}|d^* - d|/d^*$. 
% \item \textbf{Sq Rel}:$\frac{1}{|D|}\sum_{d \in D}\sqrt{|d^* - d|}/d^*$. 
% \item \textbf{RMS}: $\sqrt{\frac{1}{|D|}\sum_{d \in D}||d^* - d||^2}$. 
% \item \textbf{RMSLog}:$\sqrt{\frac{1}{|D|}\sum_{d \in D}||log(d^*) - log(d)||^2}$. 
% \item \textbf{$\delta_t$}: $\%$ of $d \in D$, s.t. $max(\frac{d^*}{d}, \frac{d}{d^*})<t$, where $t \in \{1.25, 1.25^2, 1.25^3\}$.
% \end{enumerate}

% \addlinespace
\textit{Implementation details.} 
To train ODE-CNN, we use batch size of 8, and train it from scratch, \ie random initialized weights, with 20 epochs for all experiment. We adopt Adam optimizer with $\beta_1=0.9$, $\beta_2=0.999$, and the a step-wise learning rate decay policy, which starts at 0.0002 and is reduced by 1/2 every 3 epochs. We used $L_1$ depth loss, \ie $|D - D^*|_1$, to supervise the training. For D-CSPN, we set the transformation kernel as $3$ and iteration number to be $12$. 
Finally, we train and test ODE-CNN using single NVIDIA P40 GPU, where the training cost around 2 days and testing cost around $50ms$ for each image.

\subsection{Ablation Study}
In~\tabref{tbl:quanti_res},  we present the performance gain of each module we proposed in ODE-CNN over our test set. 
Here, "SFT" is short for type of spherical feature transform layer at the end of encoder, "CSPN" stands for the type of CSPN module at the end of decoder, and "PD" refers to the view of partial dense depth map we obtained from a depth sensor. Specifically, we simulate the front view depths in OmniImage as input in all our experiments. In the following, we majorly compare the error metric of `Abs Rel` since results from others metrics follows consistently.
At the row `UResNet` and `RectNet`, we present two baseline methods introduced in OminiDepth~\cite{360D}, which produce the SoTA results over the datasets.  Specifically, `RectNet` uses rectangular filter-banks to address the horizontal distortion occurs in equirectangular projection. 
At the third row, we show the performance of our trained ODE-CNN without the proposed modules. It already significantly outperforms the `RectNet` proposed in~\cite{360D}, Abs Rel reduced by 8.5\%, which indicates the effectiveness of our adopted backbone~\cite{ma2018self}. At the fourth row, we add in the partial dense depth map at the front view, and the results are improved further, Abs Rel reduced by 2.5\%. 
Here, although the error does not improved much, we believe the generalization ability should be much better when testing over images out of the dataset.

At the rows with `IGT`, we show the performance of ODE-CNN with SFTL via inverse gnomonic (IG) projection, and compared with the results from our backbone, adding IG-SFTL reduce the Abs Rel error largely by $16.5\%$, which demonstrates that re-sampling of neighborhood is one of the key factor for depth estimation over OmniImage. At the 6th row, `DIGT` indicates adding learnable deformation to IG transform in SFTL (\secref{sec:ftl}), which further improves Abs Rel by 5\%.  

 %Due to the difference between perceptive and equirectangular projection, we introduce the proposed feature transform layer in Sec.~\ref{sec:ftl} into our network to transform our encoder feature from perceptive model to equirectangular model. We first use the inverse gnomonic transformation (IGT), as shown in the 5th row,  When we switch the IGT to learned transformation(LT), the results are further improved by $5.2\%$.

Starting at 7th row, we study the performance of the CSPN module for depth refinement, as proposed in Sec.~\ref{sec:d-cspn}. 
Here, at the row `CSPN~\cite{cheng2018cspn}`, we first evaluate its vanilla version with our backbone, which is also shown to be very effective, reducing the Abs Rel error about 15\% from 0.0624 to 0.0539. 
Then, at the row with `IG-CSPN`, we adopt the formula of IG projection as the kernel offset in CSPN, and the results are slightly improved by 1\% from 0.0539 to 0.0523. 
Next, at the row `D-CSPN`, similar with the `DIGT`, we adding learnable deformation in IG-CSPN to allow dynamic offset in its convolutional kernel, the error is further reduce another 2\% from 0.0523 to 0.0511.
Finally, we combine the most effective modules, `DIGT` and `D-CSPN`, together and adding the partial dense depth map at the front view, yielding the best results for ODE-CNN, where the Abs Rel is reduced from 0.0642 to 0.0467, or $27.3\%$ relative error reduction \wrt our baseline, and outperforms previous SoTA~\cite{360D} network by $33.4\%$.

\subsection{Qualitative Results}
In \figref{fig:offset}, we visualize the learned deformation with 3 propagtion steps inside D-CSPN at three different locations in an image, \ie at planar surface(\figref{fig:offset}(a)), within detailed structure(\figref{fig:offset}(b)) and at intersection between different surfaces(\figref{fig:offset}(c)). As can be seen, to achieve high estimation quality, on one hand, the deformation follows the pattern from inverse gnomonic projection at the planar surface to compensate the image distortion. On the other hand, it jointly considers the image content, where the context neighborhoods are more concentrated at detailed structures around the boundaries.

In \figref{fig:quali}, we illustrate two examples of estimated results from our ODE-CNN, which are compared against the results from `RectNet`. We highlight the most improved regions in the dashed rectangle, where our estimated depths not only more accurate in the estimated absolute scales, but also reveal the detailed scene structure.

% \vspace{-0.2\baselineskip}
\section{Conclusion}
% \vspace{-0.1\baselineskip}
In this paper, we introduce a novel low-cost omnidirectional depth sensing system, which combines one omnidirectioanl camera with a regular pinhole depth sensor. To our best knowledge, it provides the SoTA trade-off between dense depth sensing accuracy and efficiency comparing to other sensing options such as multi-sensor and LiDAR. 
To successfully extend the depth map with limited FoV to omnidirectional, we design ODE-CNN, where two critical convolutional modules are introduced, \ie SFTL and D-CSPN. Both modules embeds inverse gnomonic projection for handling the image distortion, and a learnable deformation for jointly consider image context. Our results achieve the SoTA performance (33\% depth error reduction vs. the baseline) over the large scale OmniDepth~\cite{360D} dataset and in the future, we will put it on device by further improve its efficiency.

\clearpage
\bibliographystyle{IEEEtran}
\bibliography{references}

\end{document}